# A Computational Approach to Measure Empathy and Theory-of-Mind from Written Texts

**Yoon Kyung Lee**[1], **Inju Lee**[1], **Jae Eun Park**[1],
**Yoonwon Jung**[1], **Jiwon Kim**[2], **Sowon Hahn**[1]

Department of Psychology, Seoul National University, Korea[1]
Chungnam National University, Korea[2]

{yoonlee78, ijlee37, dawn2089, ywjung, swhahn}@snu.ac.kr  jkim175@cnu.ac.kr

## Abstract

Theory-of-mind (ToM), a human ability to infer others' intentions and thoughts, is an essential part of empathetic experiences. We provide here the framework for using NLP models to measure ToM expressed in written texts. For this purpose, we introduce ToM-Diary, a crowdsourced 18,238 diaries with 74,014 Korean sentences annotated with different ToM levels. Each diary was annotated with ToM levels by trained psychology students and reviewed by selected psychology experts.The annotators first divided the diaries based on whether they mentioned other people: self-focused and other-focused. Examples of self-focused sentences are "I went to the mall", or "I am feeling good". The other-focused sentences were further classified into one of three levels: the writer 1) mentions the other's presence without inferring their mental state (e.g., "I saw a man walking down the street"), 2) fails to take the perspective of others (e.g., "I don't understand why they refuse to wear masks"), or 3) successfully takes the perspective of others ("It must have been hard for them to continue working")'. We tested whether state-of-the-art transformer-based models (e.g., BERT) could predict underlying ToM levels in sentences. We found that BERT more successfully detected self-focused sentences than other-focused ones. Sentences that successfully take the perspective of others (the highest ToM level) were the most difficult to predict. Our study suggests a promising direction for large-scale and computational approaches for identifying writer's ability to empathize and take perspective of others. The dataset is available at https://github.com/humanfactorspsych/covid19-tom-empathy-diary

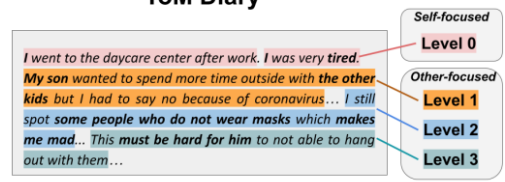

Figure 1: a diary snippet from ToM diary. Each diary can have multiple numbers of ToM levels. Sentences that did not mention other people were automatically considered as self-focused ones and labeled as Level 0. Other sentences that mentioned other's presence or thoughts were labeled according to the criteria: whether they merely mention others' presence, fail to infer other's mental states, and successfully infer other's mental states.

## 1 Introduction

The outbreak of global COVID-19 pandemic had led to an increase in expressions of one's emotions and mental health problems online (e.g., on twitter or other social media). COVID-19 pandemic is mentally stressful not only because of the fear of infection but also because of the regulatory actions and the public scrutiny that follows violation of such regulations. Large-scale text data from online are full of discussions of such subjects and therefore enables us to collect an account of the diverse perspectives and experiences under the pandemic that is hard to collect offline.

Theory-of-mind is the ability to infer another's mental state (Baron-Cohen et al., 1985; Baron-Cohen, 2000, 2001). It is one of the essential social skills that enables people to emotionally empathize with others. Lack of ToM ability has been known to negatively impact one's mental health and is associated with autism spectrum disorder, schizophrenia, and sociopathy (Brüne & Brüne-Cohrs, 2006; Goldstein & Winner, 2012). A typical child learns theory-of-mind when they are 3 or 4 years old. People can tell that a child acquired ToM when the child can understand the fact that other people may see different things from what one sees or thinks. We, therefore, used ToM as a concept as a proxy for detecting writers' intention to take the perspective of others, and hence suggest a new direction on measuring emotions and cognitive perspective-taking toward others in online communities.

Past research suggested distinguishing one's ToM into multiple levels (e.g., Level-1 and -2 perspective-taking by Flavell et al.,1981; 5 levels of visual perspective-taking by Howlin et al., 1999). Most past research measuring ToM mostly depended on using visual aids (i.e., pictures depicting others' behavior or thoughts), or survey with a small number of samples offline. Such mentioned measures, therefore, cannot be used for large-scale text corpus from online communities.

We introduce ToM-Diary, a large corpus of Korean diary collected during the COVID-19 pandemic (from Oct. 2020 to April. 2021). We quantified ToM into several levels based on the core concept of ToM: whether the writer 1) distinguishes one's own thoughts from those of others, and 2) simulates others' thoughts in their mind, going through possible reasons that can explain actions of others. We then developed a ToM annotation guideline with psychology experts based on the aforementioned criteria. We fine-tuned the state-of-the-art neural networks to our corpus for a text classification task where we assigned to sentences one of four categories based on ToM level. Then, we compared the predictions from human and state-of-the-art neural networks.

## 2 ToM-Diary

A total of 3,998 workers participated in building ToM-Diary. The number of female workers was 3,041 (76%). Mean age of the workers was 32.92 (range 18 to 88, *SD*=11.45).

### 2.1 Data Preparation

Diaries were collected a total of three stages (Figure 2). Throughout the process, we finalized an annotation guideline and reviewed the annotated outputs. Prior to diary entry, crowdworkers first read and signed the agreement form notifying them that any data collected in the project can be used or disclosed to the public for research purposes. They then filled out survey questions related to mental health: stress, loneliness, empathy, and personality and selected two emotions (out of happiness, sadness, fear, disgust, anger, surprise, calm, and 'others'). The reviewers for the crowdsourced work approved the diaries that fulfilled the requirement, then rewarded workers with monetary compensation (7,000 KRW, which is equivalent to $7 in USD). Ethical concerns related to writing personal information in diary are explained in detail in 7. Ethics section.

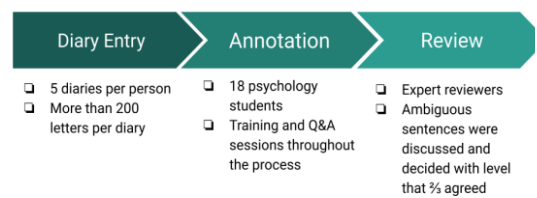

Figure 2: Overall process of data preparation. We adopted a two-step review process over the contents of the diary: the annotators of psychology students and expert reviewers.

**Stage 1. annotation guideline preparation and reviewer training**: We first classified each sentence into *self-focused* or *other-focused* based on whether other people were mentioned or not in a sentence. If there was no mention of other people, it was considered as level 0, or self-focused. If the sentence had mentions of other people (i.e., other-focused), we further distinguished them into three levels of ToM. In level 1 sentences, the writer simply described others' actions with no mentions of their thoughts or purposes; In level 2 sentences, the writer failed (or refused) to take the perspective of others by jumping to conclusions or blindly criticizing them; In level 3 sentences, the writer took the perspective of others. See Appendix A for additional details.

Reviewers consist of six psychology experts (graduate level) and made annotation guideline. The purpose of the guideline was to build a consistent set of rules for both annotators and reviewers. To ensure inter-rater agreement, three reviewers annotated the same sentence with ToM labels while practicing and compared the result to see if there was any mismatch. The inter-rater agreement was over 70% on 300 samples. Based on this practice, we then made some revisions to make the final version of a guideline.

**Stage 2. annotator recruitment and guideline training:** We recruited annotators who either held a degree in psychology from a four-year university or are currently enrolled in undergraduate university majoring in psychology. We held an online recruitment and more than 30 candidates were eligible for annotation tests. The test contained 3 questions. The first question was about defining ToM, and the other two contained practice annotation tests similar to the ones they will encounter in the actual work. We selected 30 annotators with this test, who were then required to attend an online workshop held by reviewers. Here, the annotators could learn specifics of the annotation guidelines and ask detailed questions.

**Stage 3. annotation and review:** The selected annotators labeled the diaries according to the provided guideline and were paid 150 KRW per diary (equivalent to 1.50 US cent). Annotators were to find any sentence(s) that should be labeled as level 1, level 2, or level 3. If there were no sentences in those categories, all the sentences in the diary were automatically labeled level 0. Submitted works were either approved or rejected by assigned reviewers. The rejected diaries were returned to the annotators with a detailed feedback. Once every few months, the reviewers exchanged the annotators they were in charge of so that the annotators could get balanced feedback. Also, we set up an online chatroom where all the annotators could freely ask questions about ambiguous cases where the label of the sentence seemed unclear. The reviewers answered these questions and if needed, provided some additional guidelines after further discussion.

1) The **government** extended the current social distancing level for another two weeks.

2) **They** think there is relatively low risk of infection when exercising outdoors.

3) The **protagonist of the novel** was depressed due to financial problems.

Table 1: Examples of ambiguous sentences. 1) Organizations such as the government were only considered as 'other people' when its intentions were specified, 2) It is unclear whether the writer inferred the minds of 'them,' or actually heard their opinion, in which case it was labeled as level 1. 3) Fictional characters were not considered as 'other people' because in most cases there was no need for inferring their mind, since their inner thoughts were often described in text

The reviewers had a meeting once a week in order to discuss ambiguous sentences that came up in the reviewing process (Table 1). This was to ensure that the reviewers were in accord with each other regarding the criteria for labeling these ambiguous sentences. The ambiguous sentences were reviewed and annotated with the categories that won the majority of the votes. Constant communication between and among the reviewers and the labelers allowed us to develop a consistent set of guidelines that covered diverse expressions of ToM.

## 2.2 Text Preprocessing

While level 1 to 3 was annotated in sentence-level, level 0 was initially annotated in document-level. We, therefore, we split the level-0 diaries into sentences using `KSS`[1] (Korean Sentence Splitter) library. We corrected spelling and spacing of each sentence using `py-hanspell`[2] library and removed punctuations of the sentences.

**Dealing with Imbalanced Dataset:** Among a total of 74,014 sentences, level 0 were 23,874 (32.26%), level 1 were 43,770 (59.14%), level 2 were 3,339 (4.51%), and level 3 were 3,031 (4.1%).

---

[1] https://github.com/likejazz/korean-sentence-splitter

[2] https://github.com/ssut/py-hanspell

We performed random under-sampling based on the length of sentence. For each level, we grouped the sentences by length. Above 68% of sentences in level 0 and in level 1, and above 50% of level 2 and 3 had less than 50 (Korean) letters. We randomly selected the same number of sentences as level 3 in each length group. As a result, there were 3,031 sentences in each level with a total of 12,124 sentences.

## 3 Experiment

### 3.1 Training Classifier

We trained four classifiers for predicting the underlying ToM level in a sentence from the under-sampled dataset: Multinomial Naïve Bayes (MLB; McCallum & Nigam, 1998), Feed-Forward Neural Network (FFNN; Svozil et al.,1997), Bidirectional LSTM (BiLSTM; Schuster & Paliwal, 1997), and Bidirectional Encoder Representations from Transformer (BERT)-based classifier (Devlin et al., 2018). For MLB, we employed the TF-IDF method to vectorize the sentences and used unigrams and bigrams for training the model. We employed a MNB model provided by `scikit-learn`. For FFNN and BiLSTM, we used `Khaiii`[3] (Kakao Hangul Analyzer III) for tokenizing the sentences and `Pytorch` library to process the dataset and build the models. For BERT-based classifier, we fine-tuned the pre-trained model ('bert-base-multilingual-cased') and employed the 'BertForSequenceClassification' model from the Hugging Face `transformer` library[4].

### 3.2 Training, Validation, and Test Sets of the Dataset

We split the under-sampled dataset into training, validation, and test sets with the same number of sentences for each ToM level. There were 9,700 sentences (2,425 sentences for each ToM level) for the training set and 1,212 sentences (303 sentences for each ToM level) for validation and test sets.

### 3.3 Automatic Evaluation

|  |  | MNB | FFNN | Bi LSTM | BERT |
|---|---|---|---|---|---|
|  | Precision | 0.64 | 0.56 | 0.73 | **0.78** |
|  | Recall | 0.64 | 0.56 | 0.73 | **0.78** |
|  | F1 Score | 0.64 | 0.56 | 0.73 | **0.78** |
| Acc | Level 0 | 0.63 | 0.72 | 0.85 | **0.89** |
|  | Level 1 | 0.52 | 0.47 | 0.59 | **0.76** |
|  | Level 2 | **0.83** | 0.64 | 0.75 | 0.75 |
|  | Level 3 | 0.55 | 0.42 | **0.73** | 0.72 |

Table 2: Evaluation metrics for ToM level predictions with balance dataset

Table 2 shows the performances of the models. We calculated precision, recall and F1 score for the overall performance of the models and accuracy for the performance of predicting each ToM level.

|  |  | MNB | FFNN | Bi LSTM | BERT |
|---|---|---|---|---|---|
|  | Precision | 0.29 | 0.60 | 0.77 | **0.76** |
|  | Recall | 0.29 | 0.56 | 0.72 | **0.76** |
|  | F1 Score | 0.29 | 0.51 | 0.72 | **0.76** |
| Acc | Level 0 | 0.16 | 0.83 | 0.88 | **0.94** |
|  | Level 1 | **0.99** | 0.86 | 0.92 | 0.87 |
|  | Level 2 | 0 | 0.40 | **0.63** | 0.62 |
|  | Level 3 | 0 | 0.13 | 0.46 | **0.62** |

Table 3: ToM level prediction results with imbalanced dataset

**Balanced vs Imbalanced Dataset:** Table 3 presents the model performances trained from the imbalanced dataset. We trained the classifiers using both under-sampled dataset (balanced dataset) and original dataset (imbalanced dataset). For the comparison, we split the imbalanced dataset into training, validation, and test sets. There were 71,590 sentences for the training set with 23,268 sentences for level 0, 43,164 sentences for level 1, 2,733 sentences for level 2, and 2,425 sentences for level 3. For validation and test sets, there were 1,212 sentences with 303 sentences for each ToM level. The validation and test sets were the same as that of the balanced dataset. We only used training set for MNB exceptionally.

---

[3] https://github.com/kakao/khaiii

[4] https://huggingface.co/

|  |  | All POS | Core POS |
|---|---|---|---|
|  | Precision | **0.73** | 0.65 |
|  | Recall | **0.73** | 0.66 |
|  | F1 Score | **0.73** | 0.65 |
| Acc | Level 0 | **0.85** | 0.83 |
|  | Level 1 | **0.59** | 0.53 |
|  | Level 2 | **0.75** | 0.71 |
|  | Level 3 | **0.73** | 0.56 |

Table 4: ToM level prediction results of a model trained with all POS and with core POS

**Training with All POS Tags vs with Core POS tags only:** Table 4 shows the model performances of each case. For the BiLSTM, we trained the model with all POS and only with core POS (noun, verb, adjective, and adverb), respectively.

|  |  | FFNN | BiLSTM |
|---|---|---|---|
|  | Precision | 0.50 | **0.73** |
|  | Recall | 0.50 | **0.71** |
|  | F1 Score | 0.50 | **0.71** |
| Acc | Level 0 | 0.58 | **0.84** |
|  | Level 1 | 0.44 | **0.67** |
|  | Level 2 | 0.51 | **0.57** |
|  | Level 3 | 0.46 | **0.78** |

Table 5: ToM level prediction results with the pre-trained word2vec model

**Training with vs without Pre-trained Word Embedding Model:** Table 5 shows the model performances of each case. We trained a Word2Vec (Mikolov et al., 2013) model with skip-gram model using 42,128 sentences of 19,025 diaries. For FFNN and BiLSTM, we trained the models from scratch and with the pre-trained word2vec model, respectively.

## 4 Qualitative Evaluation

Table 6 shows the results of model inference. We conducted model inference with a total of 80 test sentences. We made two different levels by difficulty: easy and difficult. Easy level sentences were similar to typical sentences of each ToM level and easy to infer the underlying ToM level. Difficult level sentences (another half) were relatively difficult to infer the underlying ToM level. We made difficult sentences similar to the ones that reviewers and annotators often disagree (e.g., whether the writer tried to take others' perspective or whether there were specific others in the sentence).

|  |  | MNB | FFNN | Bi LSTM | BERT |
|---|---|---|---|---|---|
| Easy level | Level 0 | 1.0 | 0.8 | 1.0 | **1.0** |
|  | Level 1 | 0.6 | 0.8 | 0.9 | **1.0** |
|  | Level 2 | 1.0 | 0.8 | 0.8 | **1.0** |
|  | Level 3 | 0.9 | 0.6 | 0.8 | **1.0** |
| Difficult level | Level 0 | 0.3 | 0.3 | 0.4 | **0.7** |
|  | Level 1 | 0.5 | 0.3 | 0.4 | **0.6** |
|  | Level 2 | **0.9** | 0.8 | 0.6 | 0.5 |
|  | Level 3 | 0.6 | **0.1** | 0.7 | 0.5 |

Table 6: Results of model inference

## 5 Discussion

Our results suggest that the BERT-based classifier more successfully predicted the underlying ToM level of sentence than the other three models. The BiLSTM sometimes performed better than the BERT-based classifier, for level 3 in specific. This result suggests the overall context of the sentence should be maintained to judge whether the writer tried to infer others' mental state or not. Also, the BiLSTM classifier more successfully recognized level 3 when trained with all POS. We assumed that the delicate nuances of the sentences were crucial to recognize level 3, and it can be detected by other POS except for core POS, such as postpositions and interjections. This was also apparent in annotation process as well. Detecting subtle nuances such as writer's intention to infer other people's mental state and defining the scope of 'other people' as a target of mind inference. The latter was particularly trickier because people tend to attribute mind to innate objects or beings (e.g., the virus, school, or religious figure). We, however, were able to build a consistent set of rules to quantify and distinguish ToM from text after continuous discussions among annotators and reviewers. Future research should predict ToM levels from text and posts from different online communities with different demographics of users may show different level of ToM and empathy on same issue.

## 6 Conclusion

We collected the ToM diary, a large diary corpus of daily emotional experiences (74K sentences). We introduce measures to characterize and quantify ToM expressed in written text by annotating each sentence by ToM levels. Each diary was annotated according to ToM levels by both psychology major students and expert reviewers. We found that self-focused sentences (level 0) were overwhelmingly more frequent than other-focused sentences (level 1 to 3). This was due to the fact that it was a diary (narrative story of one's daily experience). We found that transformer-based neural networks more successfully detected self-focused sentences than other-focused ones, and that level 3 sentences were the most difficult to predict. We hope our findings bring promising direction for computational approaches for exploring human cognition in large-scale.

## 7 Ethics

Upon participating in our project, crowdworkers gave informed consent. We specifically informed them of the following facts: 1) the project was funded by Data Voucher Project, Korean Data Agency; 2) any data collected can be used for research purposes and disclosed to authorized personnel. Only workers who agreed to the form voluntarily participated. In specific, due to the fact that the diary may contain personal thoughts and daily experience, we asked them NOT to include any personally identifiable information, and must remove or anonymize other people mentioned in the diary. We further took this consideration in the review process for any diary that is suspicious of disclosure of personally identifiable information. Fortunately, workers followed the guidelines by either not deliberately mentioning other people's personal information, or anonymizing them (e.g., friend A)